\documentclass[10pt, conference]{IEEEtran}
\IEEEoverridecommandlockouts
\usepackage{url}
\usepackage{cite}
\usepackage{amsmath,amssymb,amsfonts}
\usepackage[noend]{algorithmic}
\usepackage{algorithm}
\usepackage{graphicx}
\usepackage{etoolbox}
\usepackage{textcomp}
\usepackage{xcolor}
\usepackage{color}
\usepackage{booktabs}
\usepackage{tabularx}
\usepackage{tabulary}
\usepackage{adjustbox}
\usepackage{xspace}
\usepackage{amssymb}
\usepackage{amsmath}
\usepackage{amsthm}
\usepackage{float}
\usepackage{graphicx}
\usepackage{tabularx}
\usepackage{wrapfig}
\usepackage{footnote}
\usepackage{comment}
\usepackage{multirow}
\usepackage{array}
\usepackage{capt-of}
\usepackage{varwidth}
\usepackage{enumitem}
\usepackage{bm}
\usepackage{makecell}
\usepackage{fancyhdr}
\usepackage[caption=false]{subfig}
\def\BibTeX{{\rm B\kern-.05em{\sc i\kern-.025em b}\kern-.08em
    T\kern-.1667em\lower.7ex\hbox{E}\kern-.125emX}}

\makeatletter
\patchcmd{\@makecaption}
  {\scshape}
  {}
  {}
  {}
\makeatletter
\patchcmd{\@makecaption}
  {\\}
  {.\ }
  {}
  {}
\makeatother

\makeatletter
\newcommand{\removelatexerror}{\let\@latex@error\@gobble}
\makeatother

\begin{document}

\title{\LARGE Shoggoth: Towards Efficient Edge-Cloud Collaborative Real-Time Video Inference via Adaptive Online Learning\vspace{-0.5em}}

\author{
    \IEEEauthorblockN{
    {\normalsize Liang Wang\IEEEauthorrefmark{1}\textsuperscript{,}\IEEEauthorrefmark{2}\textsuperscript{,}\IEEEauthorrefmark{3}\thanks{\IEEEauthorrefmark{3}These two authors have contributed to this work equally.}, 
    Kai Lu\IEEEauthorrefmark{1}\textsuperscript{,}\IEEEauthorrefmark{3}, 
    Nan Zhang\IEEEauthorrefmark{2}, 
    Xiaoyang Qu\IEEEauthorrefmark{2}\IEEEauthorrefmark{4}\thanks{\IEEEauthorrefmark{4}Corresponding authors.}, 
    Jianzong Wang\IEEEauthorrefmark{2},
    Jiguang Wan\IEEEauthorrefmark{1}\IEEEauthorrefmark{4}, 
    Guokuan Li\IEEEauthorrefmark{1}, 
    Jing Xiao\IEEEauthorrefmark{2}}}
    \IEEEauthorblockA{{\small \IEEEauthorrefmark{1}Huazhong University of Science and Technology, China}
    \\\{\small {iggiewang, emperorlu, jgwan, liguokuan\}@hust.edu.cn}}
    \IEEEauthorblockA{{\small \IEEEauthorrefmark{2}Ping An Technology (Shenzhen) Co., Ltd., China}
    \\\{\small {nzhang889@gmail.com, quxiaoy@gmail.com, jzwang@188.com, xiaojing661@pingan.com.cn\}}}
    \vspace{-2.2em}
}

\maketitle

\begin{abstract}

This paper proposes Shoggoth, an efficient edge-cloud collaborative architecture, for boosting inference performance on real-time video of changing scenes. Shoggoth uses online knowledge distillation to improve the accuracy of models suffering from data drift and offloads the labeling process to the cloud, alleviating constrained resources of edge devices. At the edge, we design adaptive training using small batches to adapt models under limited computing power, and adaptive sampling of training frames for robustness and reducing bandwidth. The evaluations on the realistic dataset show 15\%–20\% model accuracy improvement compared to the edge-only strategy and fewer network costs than the cloud-only strategy.

\end{abstract}

\section{Introduction}
\vskip -2pt

Real-time video inference, such as object detection\cite{zhao2021neural}, is a foundational component in many applications, including intelligent traffic surveillance and automatic driving. Real-time video inference applications prefer edge devices for analytics because they require prompt feedback. Edge computing eliminates the need for costly network connections to stream videos to the cloud, which can solve high transmission latency.

Although the rapid progress in deep neural networks (DNNs) has driven the development of real-time video inference, as edge devices are resource-constrained, only DNNs with fewer weights and shallower architectures can be deployed. These lightweight DNNs are vulnerable to \textit{data drift}, i.e., real-time video data drastically varies\cite{maltoni2019continuous}. It is difficult to achieve the desired accuracy for multiple environments in one offline training, especially due to the impact of even minor variations on accuracy. However, video scenarios also change over time for a given device, e.g., crowd densities, weather, and illumination. As a result, lightweight DNN models have poor adaptability and only perform well in limited scenarios.

\begin{figure}[htbp]
\begin{center}
\includegraphics[width=0.45\textwidth]{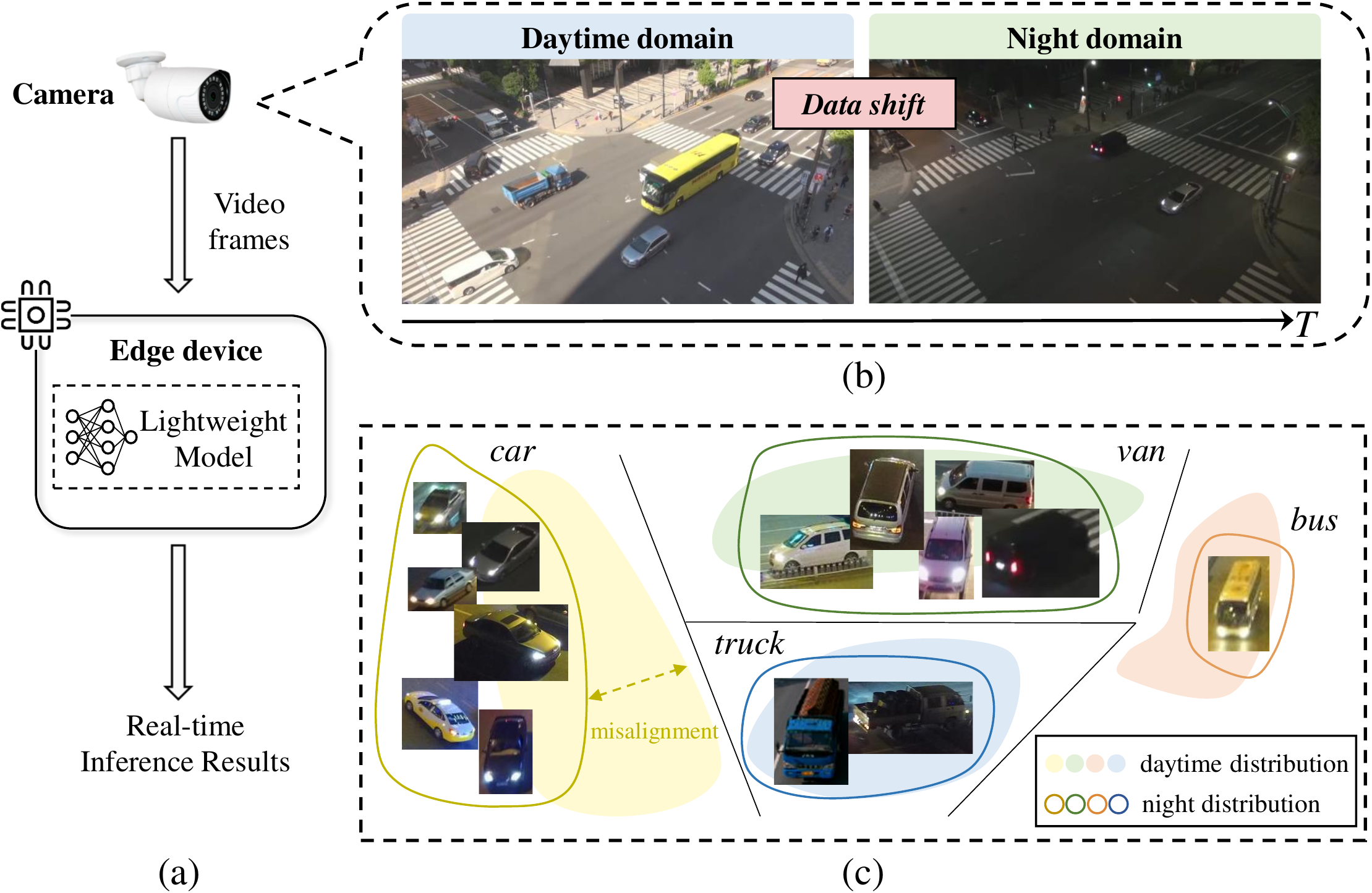}
\vskip -4pt
\caption{Illustration of \textit{data drift}. (a) displays the camera connect to the edge device for real-time video inference, (b) shows data drift from daytime to night domains due to variations in data distribution and illumination, and (c) shows the shift in class distributions of data drift, resulting in many objects at night being difficult to distinguish for the lightweight model.}
\label{fig1}
\end{center}
\vskip -16pt
\end{figure}

We argue that real-time video scenarios exhibit a significant class imbalance, and the visual appearance of objects from the same class can vary due to changes in the scene. It is the primary cause of data drift during video analysis, as shown in Figure \ref{fig1}. The strong correlation of video frames over short time intervals and the data distribution shift over long periods are notable difficulties. Thus, we must consider the ability of models to migrate from domain to domain.

One promising approach to improve inference efficiency is knowledge distillation\cite{hinton2015distilling, qu2021enhancing}. The primary idea is to prepare a complex DNN model as the teacher model and a specialized lightweight model as the student model, and continually distill knowledge from the teacher to the student. However, due to the high computational requirement of the knowledge distillation process, it is equally impossible to perform the entire knowledge distillation process online on edge devices. In fact, edge-cloud collaboration has become a prevailing computing paradigm that can effectively leverage cloud computing and edge computing\cite{li2021appealnet, gao2021intelligent, liu2021petri, liu2022sniper}. Cloud computing has higher computational efficiency, while edge computing can provide low latency. Intensive collaboration can maximize the performance of real-time video inference applications.

In this paper, we propose \textbf{Shoggoth}, a novel edge-cloud collaborative architecture, continually adapting lightweight models running on edge devices by knowledge distillation for inference of changing videos in real-time. During performing inference, edge devices periodically send video frame samples to the cloud for online labeling to fine-tune the local model.

However, achieving such an adaptive video inference architecture based on edge-cloud collaboration is not without challenges. The first is the retraining efficiency on the device with limited computing resources. Collecting all video data accumulated throughout the lifetime of the deployed framework and retraining the entire model from scratch is infeasible, particularly given the frequent real-time changes. Conversely, solely using the newly available data to retrain the prediction model leads to catastrophic forgetting\cite{mccloskey1989catastrophic}. We demonstrate that utilizing small batches for adaptive training with replay memory is a flexible solution. The second is the communication overhead. We show that the adaptive frame sampling method over a carefully selected recent frame horizon presents robust rationality — neither too small to retrain frequently and overfit nor too big to exceed the model generalization capacity.

We evaluate Shoggoth on the real-time video object detection task with the lightweight model (YOLOv4\cite{bochkovskiy2020yolov4} with Resnet18\cite{he2016deep} backbone). Our experiments use the realistic dataset, which has different conditions like sunny, cloudy, rainy, and night. Our results illustrate the superiority in every respect of Shoggoth combined with adaptive online learning compared to edge-only, cloud-only, or other strategies.

In general, we make the following contributions:

\begin{itemize}
    \item We propose Shoggoth, an efficient real-time video inference edge-cloud collaborative architecture with adaptive online learning that addresses the data drift problem by decoupled knowledge distillation.
    \item We design adaptive training to fine-tune edge models using replay memory to combat catastrophic forgetting and adapt the limited computing power of edge devices.
    \item We design adaptive frame sampling to increase system robustness and reduce bandwidth.
\end{itemize}

\section{Related Work}

\textbf{Edge-cloud collaborative video inference.} Existing video inference solutions using edge-cloud collaboration tradeoff between low latency and high accuracy.
PETRI\cite{liu2021petri} implements a latency-hiding pipeline workflow and employs a retro-tracking method to detect missed targets at the edge, thereby reducing bandwidth to the cloud.
NoScope\cite{kang2017noscope} accelerates inference by cascading models and filtering.
Clownfish\cite{nigade2020clownfish} fuses current inference results from the specialized model running on the edge and deferred results from the complex model executing in the cloud.
Unfortunately, these solutions use offline-trained models, ignoring model adaptation and data drift on model accuracy.
In addition, Ekya\cite{bhardwaj2022ekya} and AMS\cite{khani2021real} make an initial exploration of using knowledge distillation to continually train and adapt edge models, boosting their performance on the live video. They perform both labeling and training processes on the remote server, and streaming model updates to edge devices aggravates the bandwidth consumption. In addition, allocating additional GPU time to training reduces the number of scalable edge devices supported by a single GPU.

\textbf{Data Drift.} Edge DNNs have a restricted capacity to memorize scenarios and object appearances. Therefore, they are especially susceptible to data drift\cite{maltoni2019continuous}, which occurs when real-time video data diverges significantly from domain to domain. Temporal fluctuations in scene density (e.g., rush hour) and illumination (e.g., daytime vs. nighttime, sunny vs. rainy) present a significant challenge for traffic cameras to achieve accurate object detection. In addition, the data distribution of objects changes over time, which decreases the accuracy of the edge model\cite{liu2019large}. Due to their limited capacity to remember variations, edge DNNs must be continually fine-tuned with recent video data to achieve high accuracy.

\textbf{Online continual learning.} Continual learning (CL) has gained increasing interest recently in computer vision\cite{aljundi2019gradient, pellegrini2020latent}. We investigate a more realistic and difficult situation in which data are delivered in small batches, and models are trained once on each batch. Our strategy depends on representative memory replays\cite{pellegrini2020latent}. As video streams arrive endlessly in real-time, it is laborious to annotate all the video frames for computational training models. Reducing the burdensome costs of labeling remains an under-explored and challenging problem in CL. In our case, online distillation and adaptive frame sampling are applied to reduce the onerous costs of labeling. Overall, we design adaptive online learning, focusing on training with small online-labeled batches compatible with the limited computing resources available on edge devices.

\section{Shoggoth Design}

\vskip -4pt
\subsection{System Overview}
\vskip -2pt

\begin{figure}[htbp]
\begin{center}
\vskip -6pt
 \includegraphics[width=0.47\textwidth]{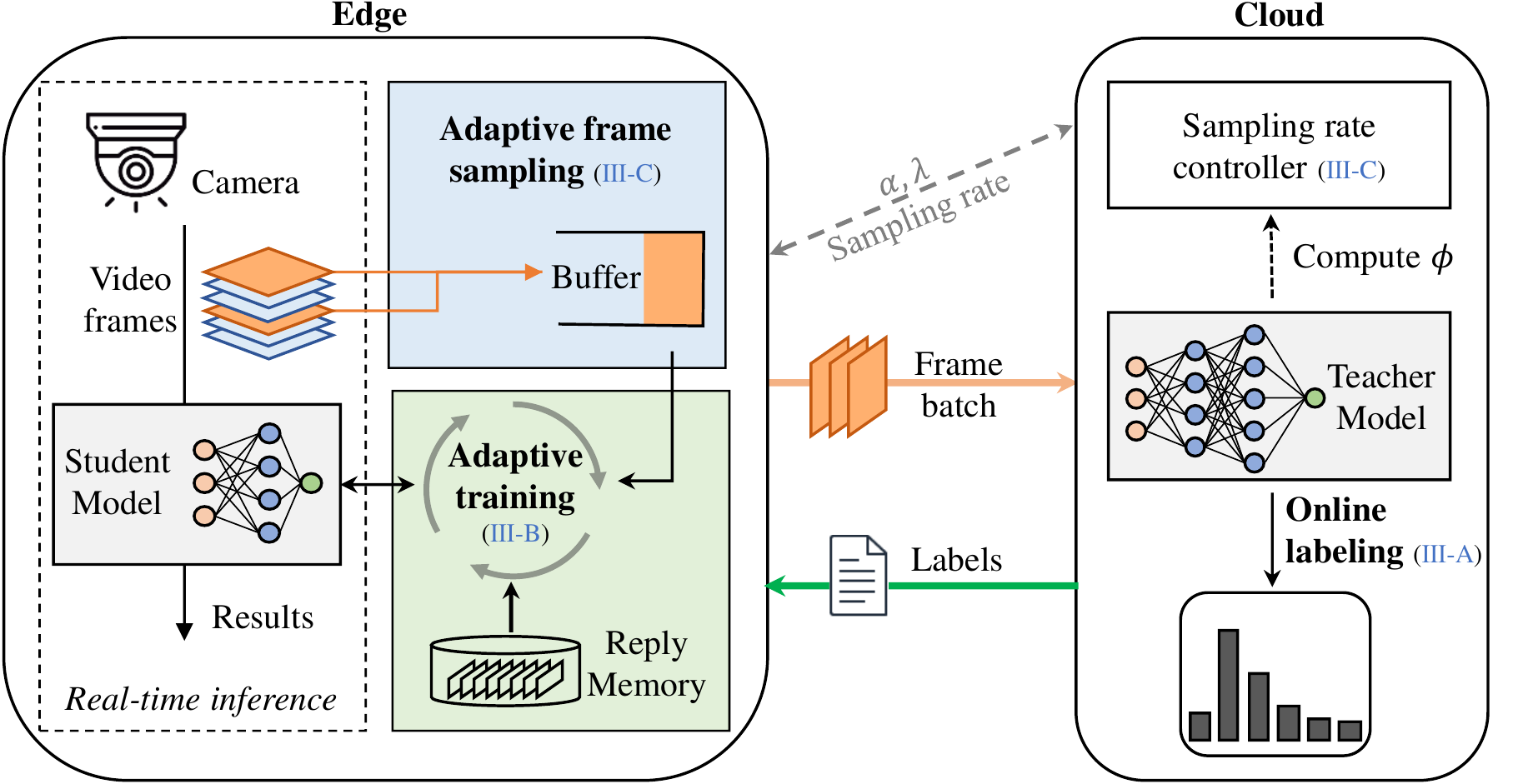}
\end{center}
\vskip -8pt
\caption{Shoggoth system overview.}
\label{shoggoth}
\vskip -4pt
\end{figure}

Figure \ref{shoggoth} shows the overall architecture of Shoggoth, comprising two parts: the cloud and the edge. The ultimate goal of this architecture is to achieve high real-time video inference efficiency no matter what changing scenarios.

Real-time video inference runs at the edge, receiving all video frames and outputting the results. \textbf{Decoupled knowledge distillation} is designed in Shoggoth architecture, which decouples the labeling process and the training process, offloading the labeling process to the cloud and performing the training process at the edge device. The cloud communicates with the edge over the network. All edge devices share the complex computation-intensive teacher model in the cloud, which has been pre-trained on extensive image datasets and maneuvers billions of model parameters. With abundant computing resources in the cloud, the teacher model is capable of running on the cloud server with high accuracy. Edge devices periodically collect video frame samples and send them to the cloud. The cloud server labels them online and sends them with labels back. Subsequently, the edge device utilizes them to fine-tune the lightweight student model, improving the accuracy of real-time video inference for the current environment. The entire fine-tuning workflow is described as \textbf{adaptive online learning}, which is divided into two stages, online labeling and adaptive training.

\textbf{Online labeling.} The cloud server labels the incoming video frames online by the teacher model detector. Shoggoth naively treats pseudo-labeled data from various domains and labeled data equally for loss. All pseudo-labeled samples are assigned the label of 1, i.e., for the $i$-th training sample $X_i$ (not an image, but a region in an image), the label $y_i$ is defined as

\vskip -8pt
\begin{equation}
y_i= \begin{cases}1, & \text { if } X_i \text { is a pos. sample (from detector). } \\ 0, & \text { if } X_i \text { is a neg. sample. }\end{cases}  \label{yi}
\end{equation}

\textbf{Adaptive training.} The edge device trains the student model to adapt for scene changes and minimize the loss over the samples from the buffer in the current video. Specifically, we design \textbf{adaptive training with replay memory} to meet resource-constrained edge devices' computational and memory requirements, which is detailed in Section \ref{adaptive training}.

\textbf{Optimization}. Practically, the cloud server should dynamically adapt the frame sampling rate utilized by the edge device, depending on the video characteristics (how quickly scenarios change). To \textbf{increase robustness and reduce bandwidth}, the adaptive frame sampling algorithm is designed to modify each device's frame sampling rate via the sample rate controller in the cloud. This is discussed in Section \ref{adaptive sampling}. To be specific, we describe our design using the object detection task as an example, but the architecture is general and able to be extended to various video inference applications.

\subsection{Adaptive Training with Replay Memory}  \label{adaptive training}
\vskip -2pt

Adaptive training aims to adjust the model to handle domain change scenarios. Through adaptive training, the model continually accumulates knowledge based on new data and improves its performance. The main challenge is catastrophic forgetting\cite{mccloskey1989catastrophic}, which denotes the model forgetting data learned in the past. Replay is a widely recognized approach to address the catastrophic forgetting issue.

\begin{figure}[htbp]
\begin{center}
 \includegraphics[width=0.47\textwidth]{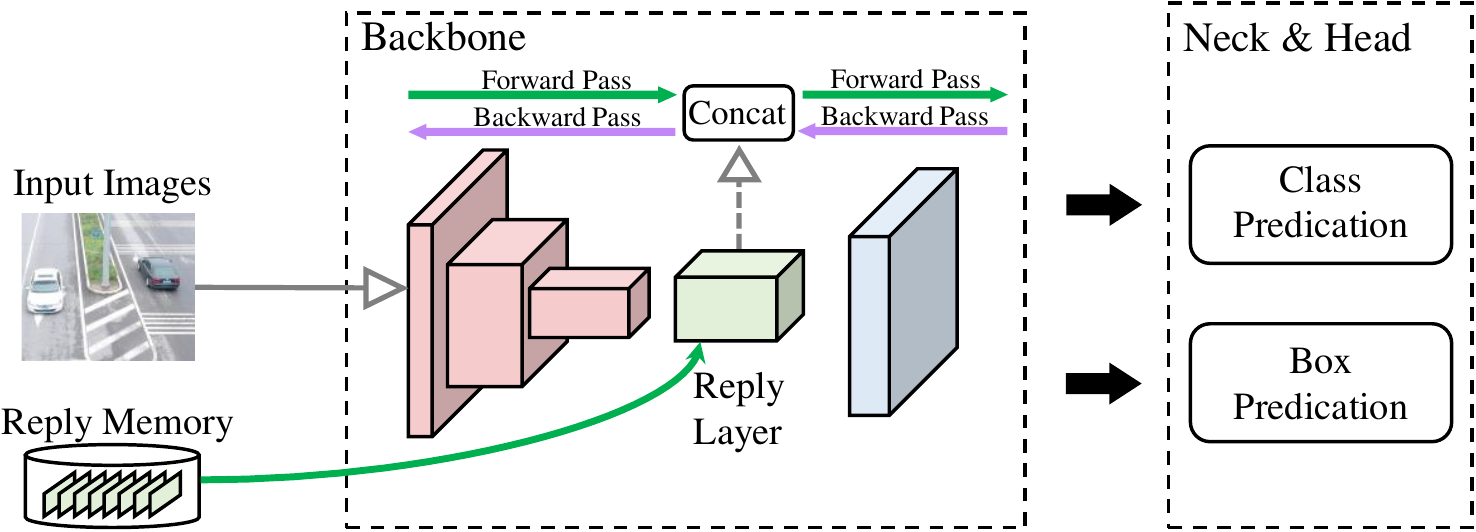}
\end{center}
\vskip -8pt
\caption{Adaptive Training Schema of Object Detector with Replay Memory.}
\label{replay-memory}
\vskip -4pt
\end{figure}

However, adaptive training on the resource-constrained edge device poses some questions on both the efficiency and sustainability of the process. We implement a lightweight strategy for real-time on-device adaptation without forgetting. 
Our solution focuses on addressing the relevant data drift in real-time video over time and meeting the computational and memory requirements of edge devices. 
The proposed solution applied to object detection paradigm can be seen in Figure \ref{replay-memory}.

Forgetting is mainly localized at the classification head, which refers to the last fully connected layer of the adopted network, and its tuning is essential to maximize accuracy. The layers near the input are responsible for low-level feature extraction, and the weights are pretty stable and reusable in different scenes after adequate pre-training. On this basis, the replay memory is designed to store the activation volumes of images in a specific layer (Replay Layer) instead of raw input image data. To keep the validity of the stored activations, the training process is supposed to decrease the learning rate of all layers before the replay layer and allow full learning of all layers after the replay layer. This technique can speed up the training process on a pre-trained model.

\begin{figure}
\renewcommand{\algorithmicrequire}{\textbf{Input:}}
\renewcommand{\algorithmicensure}{\textbf{Output:}}
\removelatexerror
\vskip -12pt
\begin{algorithm}[H]
\caption{Replay Memory Management}
\begin{algorithmic}[1]
    \STATE $\mathcal{M} \leftarrow \varnothing$
    \STATE $\mathcal{M}_{\text{size}} \leftarrow \text{number of images memorized in } \mathcal{M}$
    \FOR{\textbf{each} adaptive training {\it i}}
    \STATE $\mathcal{B} \leftarrow \text{current training batch}$
    \STATE train the model on $\mathcal{B} \cup \mathcal{M}$
    \IF{\textsc{Is}\textsc{Full}($\mathcal{M}$)}
    \STATE $h \leftarrow \frac{\mathcal{M}_{\text{size}}}{i}$
    \STATE $\mathcal{M}_{\text{add}} \leftarrow \text{random sampling of } h \text { images from } \mathcal{B}$
    \STATE $\mathcal{M}_{\text{replace}} \leftarrow \text{random sampling of } h \text{ images from } \mathcal{M} $
    \STATE $\mathcal{M} \leftarrow \left(\mathcal{M}-\mathcal{M}_{\text{replace}}\right) \cup \mathcal{M}_{\text{add}}$
    \ELSE
    \STATE $\mathcal{M} \leftarrow \mathcal{M} \cup \mathcal{M}_{\text{add}}$
    \ENDIF
    \STATE \textbf{reset} $\mathcal{B}$
    \ENDFOR
\end{algorithmic}
\end{algorithm}
\label{alg1}
\vskip -20pt
\end{figure}

Adaptive training utilizes mini-batch SGD: (i) the forward step involves a concatenation at the replay layer that combines images across the front layers and activations from the replay memory; (ii) the backward step updates the weights. In the extreme case that the front layers are entirely frozen (i.e., decelerate to 0), the backward pass is terminated just before the replay layer for replay memory. However, in ordinary cases where the front layers are not entirely frozen, replay memory activations are affected by an aging effect, meaning they increasingly deviate over time from the activations that the same image would generate if it is fed from the input layer. Nevertheless, if the front layer training is slow enough, the aging impact is negligible since the replay memory has sufficient time to be updated with new images.

\textbf{Replay Memory Management.} Replay memory allows data augmentation to work properly, and managing replay memory is a crucial aspect of adaptive training. Algorithm 1 illustrates replay memory management. Replay memory $\mathcal{M}$ updates are triggered only after an adaptive training run. During each retraining batch $\mathcal{B}$, a randomly selected subset of images from the batch replaces an equally random subset of images in the replay memory, keeping the replay memory at the proper size. In particular, if replay memory is not full (during the initial runs), all the available images are memorized. Finally, the current batch is emptied for preparing the next training. The replay memory management ensures that each training batch sampled during a training session has an equal probability of being stored in the replay memory, thus improving the model's learning and preventing forgetting.

\textbf{Training Control.} A constant proportion of original and replay images is maintained to streamline training. If every training batch contains $N$ images and the replay memory includes $M$ images, within a mini-batch of size $K$ we concatenate $\frac{K \times N}{N + M}$ original images (of the current batch) with $\frac{K \times M}{N + M}$ replay images, i.e., only $\frac{K \times N}{N + M}$ images need to cross the red layers of Figure \ref{replay-memory}. An efficient strategy to implement learning slowdown in the front layers is to freeze the weights by adjusting the learning rate to 0 after first batch, while making the batch normalization (BN) moments adapt freely to the input image statistics across all batches. Further, in the adopted model, BN layers are replaced with Batch Renormalization (BRN) \cite{ioffe2017batch} layers, which has been shown to be an effective way of controlling internal covariate shift, hence making learning with fine-grained batches faster and more robust.

\subsection{Increasing Robustness and Reducing Bandwidth}  \label{adaptive sampling}
\vskip -2pt

Models for videos where the scene is constantly changing require frequent retraining, while models for stationary videos with little scene change only need to be retrained at longer intervals. Thus, adaptive training frequency on different edge devices is supposed to be different and changing for robustness to handle scene variations better—the model will not overfit narrowly or surpass the generalization capacity. Meanwhile, frequency is affected by the frame sampling rate because each edge device uploads a certain number of frames at a time to the cloud for labeling and then performs adaptive training. We design the adaptive frame sampling approach, dynamically adjusting the frame sampling rate on edge devices depending on the speed and degree of scene changes observed in videos, inference accuracy, and resource usage. It also reduces the edge device load for slowly-changing or stationary videos and the uplink bandwidth (edge-to-server network costs). We define three metrics to adapt the frame sampling rate:

\begin{itemize}[leftmargin=*]
\item \textbf{$\bm{\phi}$: the rate of change over time for video frames.} 
Compared with raw pixels, labels are commonly picked up in a much smaller space (e.g., a few object classes), making them a more robust signal for the measurement of change. We follow Khani \textit{et al.} \cite{khani2021real} to use labels of the teacher model to compute the $\phi$ in the cloud server.
Consider a series of frames $\left\{\mathbf{I}_k\right\}_{k=0}^n$, with $\left\{\mathcal{T}\left(\mathbf{I}_k\right)\right\}_{k=0}^n$ denoting the teacher model's output on these frames. For each frame $\mathbf{I}_k$, we define $\phi_k$ using the same loss function that is used to define the task, with $\mathcal{T}(\mathbf{I}_k)$ and $\mathcal{T}(\mathbf{I}_{k-1})$ serving as the prediction and ground-truth labels, respectively. It means $\phi_k$ is the loss (error) of the teacher model's prediction on $\mathbf{I}_k$ relative to the label $\mathcal{T}(\mathbf{I}_{k-1})$. Therefore, the smaller $\phi_k$ is, the more similar are the labels for $\mathbf{I}_k$ and $\mathbf{I}_{k-1}$, i.e., slowly-changing or stationary scenes tend to obtain lower scores.

\item \textbf{$\bm{\alpha}$: the estimated inference accuracy.} The model infers every unlabeled frame of videos in the current domain. A prediction is considered accurate if the (normalized) confidence score for the $i$-th prediction (i.e., the model’s posterior), $d_i$, exceeds a threshold $\theta$. In object detection, a commonly used value for $\theta$ is 0.5. The percentage of accurate predictions, $\alpha$, is calculated as the estimated average accuracy of video inference.

\item \textbf{$\bm{\lambda}$: the resource usage over a period of time.} Edge devices continuously collect resource usage and send the usage to the cloud. In our implementation, the target resource is simplified — only GPU or CPU resource usage in percent for every second is monitored. And a few configuration variables are provided, such as collecting frequencies.

\end{itemize}

The cloud server periodically calculates the average $\phi$ on recent frames, and collects $\alpha$ since the last adaptive training and GPU/CPU usage $\lambda$ for adaptive training from the edge device. Then, the sampling rate of the edge device is adjusted by the sampling rate controller to keep the $\phi$ close to the target value $\phi_{target}$ and the $\alpha$ near another target value $\alpha_{target}$:


\vskip -8pt
\begin{equation}
r_{t+1}=[R(\phi)+R(\alpha)+R(\lambda)]_{r_{\min }}^{r_{\max }}
\end{equation}

$R(\phi)$, $R(\alpha)$ and $R(\lambda)$ are denoted as

\vskip -4pt
\begin{equation}
\begin{aligned}
&R(\phi)=\eta_r \cdot(\bar{\phi}_t-\phi_{target}) \\
&R(\alpha)={\eta_\alpha}\cdot\max(0, \alpha_{target}-\alpha_t) \\
&R(\lambda)=(1+\bar{\lambda}_{t+1}-\bar{\lambda}_t)\cdot r_t
\end{aligned}
\end{equation}

\noindent where $\eta_r$ and $\eta_\alpha$ are the step size parameters, and the symbol $[\cdot]_{r_{\min }}^{r_{\max }}$ indicates that the sampling rate is constrained within the range $\left[r_{\min }, r_{\max }\right]$. In our case, $r_{min}$ = 0.1 frames per second (fps) and $r_{max}$ = 2 fps are adopted.

To further reduce bandwidth, the edge device buffers samples and applies H.264 video encoding standard to compact this buffer before transmission. In our experiments, compressing the buffered samples takes 1-3 seconds.

\section{Performance Evaluation}   \label{Evaluation}

\vskip -2pt
\subsection{Setup \& Methodology}
\vskip -2pt

We choose the object detection task in real-time video inference as our evaluated workload for Shoggoth, and all experiments are conducted with this task. 

\noindent \textbf{Datasets.} We evaluate Shoggoth on three typical benchmarks: UA-DETRAC\cite{wen2020ua}, KITTI (Car only)\cite{geiger2013vision} and Waymo Open\cite{sun2020scalability}. They include various kinds of weather (i.e., sunny, cloudy, and rainy) and illumination circumstances (night and day). 
UA-DETRAC offers 100 challenging video sequences in real-world traffic scenarios, consisting of more than 140,000 frames, from which we select several video sequences to link together to form a long video stream. In addition, since the KITTI and Waymo datasets do not contain sequential timestamps, we concatenated images from the same camera to create lengthy video sequences in chronological order for these datasets. Videos playback at 30 frames per second (fps). With these videos, the efficiency and effectiveness of our system under changing scenarios are evaluated.

\noindent \textbf{DNN models.} At edge devices, we use the YOLOv4\cite{bochkovskiy2020yolov4} with Resnet18\cite{he2016deep} backbone.
At the cloud server, we use an expensive golden model (Mask R-CNN\cite{he2017mask} with ResNeXt-101) to obtain ground truth labels, and we verify that the generated labels are very similar to human-annotated labels.

\begin{table*}[htbp]
\caption{Comparison of different strategies on three datasets.}
\label{table:table1}
\centering
\vskip -8pt
\setlength\tabcolsep{4pt}
\small
\begin{tabular}{l l c c c c c}
\hline
\textbf{Dataset} & \textbf{Metric} & \textbf{Edge-Only} & \textbf{Cloud-Only} & \textbf{Prompt} & \textbf{AMS} & \textbf{Shoggoth}\\
\hline
\hline
\multirow{2}{*}{\textbf{UA-DETRAC\cite{wen2020ua}~}} & Up/Down Bandwidth (Kbps) & 0/0 & 3257/3539 & 303/22 & 151/226 & 135/10 \\
& mAP@0.5 (\%) & 34.2 & 58.9 & 48.3 & 51.6 & 53.5\\
\hline
\multirow{2}{*}{\textbf{KITTI\cite{geiger2013vision}~}} & Up/Down Bandwidth (Kbps) & 0/0 & 2184/2437 & 179/10 & 94/203 & 91/5 \\
& mAP@0.5 (\%) & 56.8 & 78.0 & 71.4 & 72.8 & 74.7\\
\hline
\multirow{2}{*}{\textbf{Waymo Open\cite{sun2020scalability}~}} & Up/Down Bandwidth (Kbps) & 0/0 & 2687/2880 & 278/15 & 127/207 & 112/8 \\
& mAP@0.5 (\%) & 47.5 & 64.7 & 61.5 & 59.1 & 61.9\\
\hline
\end{tabular}
\vskip -12pt
\end{table*}

\noindent \textbf{Platforms.} We adopt NVIDIA Jetson TX2 as our edge device due to its low-power GPU and suitability for large-scale edge deployment. A cloud server equipped single NVIDIA V100 GPU is used for all evaluations.

\noindent \textbf{Strategies.} We evaluate the following strategies:
\textit{1) Shoggoth.} The edge-cloud collaborative strategy with adaptive online learning is proposed in this paper. For experiments of Shoggoth, all images are resized to 512 × 512. Every training batch contains 300 images with 1500 replay images, and the mini-batch size is 64. Replay occurs on the penultimate layer (\textit{pool}). A training session consists of 8 epochs.
\textit{2) Edge-Only.} The edge model without video-specific customization performs all inferences on the edge device. 
\textit{3) Cloud-Only.} All frames are uploaded to the cloud using the complex model for detection and returning the results. 
\textit{4) Prompt.} It can be seen as Shoggoth without adaptive sampling. The main difference from Shoggoth is that the sampling rate is configured as 2 fps, the same as the maximum frame sampling rate used in Shoggoth. Model adaptation happens promptly and regularly.
\textit{5) Adaptive Model Streaming (AMS).} Compared to Shoggoth, AMS\cite{khani2021real} performs the entire knowledge distillation in the cloud for model adaption, and the updated model is sent back to the edge device. All other configurations are the same as Shoggoth, including adaptive sampling, etc.

\noindent \textbf{Performance Metrics.} The results are mainly evaluated by uplink and downlink bandwidth and mAP@0.5 (mean Average Precision, and Intersection over Union = 0.5).

\subsection{Evaluation Results}
\vskip -2pt

Our results show the performance advantages of Shoggoth, and the impact of adaptive training and adaptive sampling.

\begin{figure}
\begin{center}
 \includegraphics[width=0.48\textwidth]{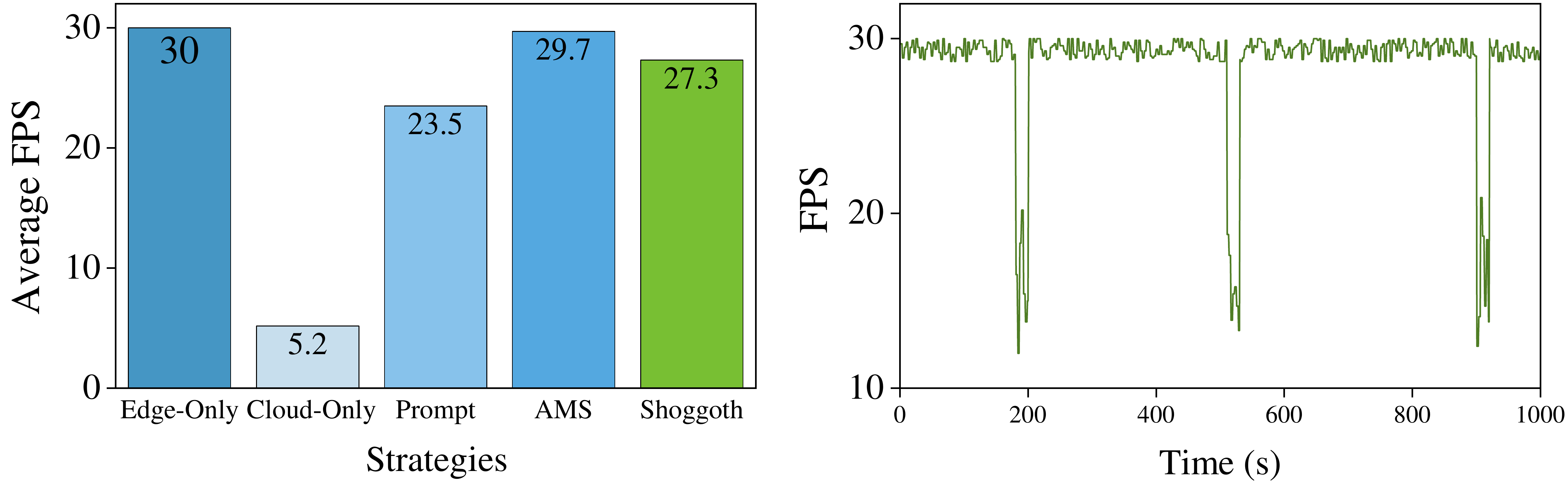}
\end{center}
\vskip -12pt
\caption{Average FPS overall for different strategies (left) and FPS over time in Shoggoth for UA-DETRAC video sequence of the initial moment (right).}
\label{e1}
\vskip -16pt
\end{figure}

\noindent \textbf{Overall improvements.} Table \ref{table:table1} summarizes the results on the three streams. The main takeaways are:

\begin{itemize}[leftmargin=12pt]
  \item [1.] 
  Shoggoth provides significant mAP score gains, achieving 15-20\% better than Edge-Only (no model adaption). Besides, Shoggoth uses significantly less network bandwidth than other strategies using the cloud.
  \item [2.]
  For Cloud-Only, all frames need to be uploaded to the cloud server for inference. Cloud-Only offers the optimum mAP accuracy, but it requires approximately 24× the uplink bandwidth and 350× the downlink bandwidth of Shoggoth. In comparison, Shoggoth brings less than 5\% degradation in mAP but saves enormous bandwidth.
  \item [3.]
  The prompt strategy incurs more bandwidth overhead, but there is no guarantee that overall model accuracy will necessarily be higher than Shoggoth with adaptive frame sampling. On videos that vary significantly over time, updating the model promptly and regularly can lead to accuracy improvement. However, customizing the model for the intervals can backfire when dealing with stationary videos with little scene change, in contrast to adaptive frame sampling, which constantly adapts the model to video content and improves accuracy.

\begin{table}
\vskip -4pt
\caption{mAP (\%) and training time (in seconds) of different methods.}
\label{table:table2}
\centering
\vskip -8pt
\setlength\tabcolsep{4pt}
\small
\begin{tabular}{c c c c c}
\hline
\multirow{2}{*}{\textbf{Method}} & \multirow{2}{*}{\textbf{mAP}} & \multicolumn{3}{c}{\textbf{Training Time}} \\
\cmidrule(lr){3-5}
 & & \textbf{Forward} & \textbf{Backward} & \textbf{Overall}\\
\hline
\hline
\multirow{1}{*}{\textit{Ours (Baseline)}} & 53.5 & 17.8 & 0.8 & 18.6 \\
\hline
Input & 49.6 & 536.2 & 31.6 & 567.8 \\
Completely Freezing & 50.7 & 17.8 & 0.7 & 18.5 \\
Conv5\_4 & 52.3 & 20.2 & 5.8 & 26.0 \\
No Replay Memory & 45.6 & 95.7 & 6.2 & 101.9 \\
\hline
\end{tabular}
\vskip -16pt
\end{table}

  \item [4.]
  AMS fine-tunes (a copy of) the edge device's model entirely leveraging sample frames on the cloud server to mimic the teacher model. It brings a similar increase in accuracy to Shoggoth. However, the need to send the updated student model leads to redundant downlink bandwidth. In addition, AMS requires more computing resources for training on the cloud, so Shoggoth can support more edge devices when several edge devices share the same GPU server.
\end{itemize}

\noindent \textbf{Impact of adaptive training.} Figure \ref{e1} shows how adaptive training affects inference. Compared to the Edge-Only, Shoggoth causes an average loss of 2.7 fps. This is because when adaptive training takes resources away from the inference, fps drops dramatically for the inference (from 30 to 15). However, the adaptive training process is fast, so we only observe a slight average loss. Then, we additionally perform an ablation study to investigate the effects of key choices in the adaptive training approach with the replay memory implementation. Table \ref{table:table2} shows mAP and training time results. We comparatively evaluate (\textit{i}) placing the replay memory on the input layer rather than the replay layer; (\textit{ii}) the front layers are completely frozen; (\textit{iii}) the \textit{conv5\_4} layer variant identifies the replay memory; (\textit{iv}) no replay memory, only the current new batch of data is used for training. Our findings validate that the use of replay memory significantly improves the system's overall performance, with the highest mAP score and almost the same training time as freezing all front layers.

\begin{figure}
\begin{center}
 \includegraphics[width=0.47\textwidth]{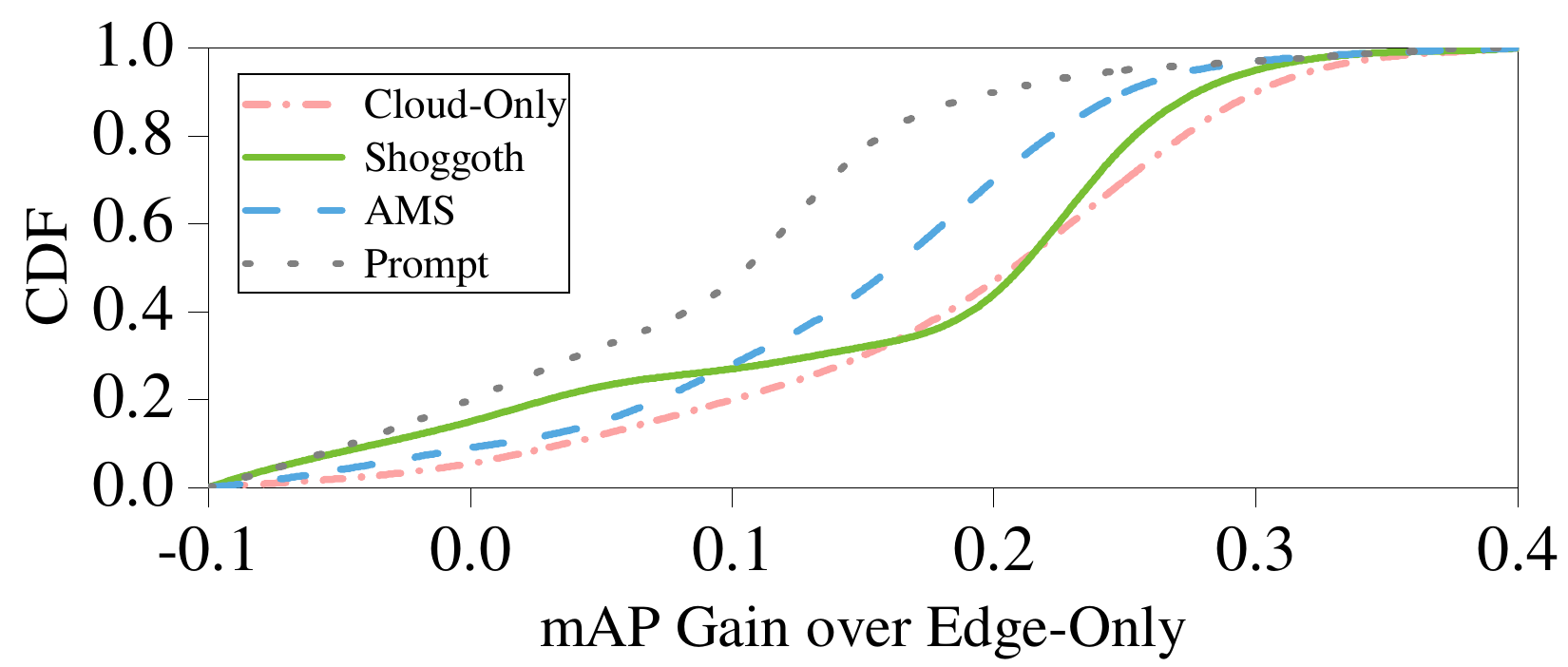}
\end{center}
\vskip -16pt
\caption{CDF of mAP gain vs. Edge-Only across all frames for other strategies.}
\label{e2}
\vskip -16pt
\end{figure}

\noindent \textbf{Impact of adaptive frame sampling.} Table \ref{table:table3} demonstrates the effect of different sampling rates on the uplink bandwidth and the average IoU of inference. No matter how much of a fixed sampling rate is adopted, it is not on a par with adaptive frame sampling in terms of accuracy (high sampling rates cause overfitting to a few recent frames). Furthermore, we compared the cumulative distribution of mAP improvement for all solutions in comparison to Edge-Only over all frames to demonstrate that Shoggoth with adaptive frame sampling shows good robustness on the overall frames in the scene rather than being limited to certain still scene segments in the video. The results are shown in Figure \ref{e2}. As expected, due to the highly accurate model and the high computational power in the cloud, Cloud-Only performs best. Even so, Shoggoth with adaptive frame sampling still slightly outperforms Cloud-Only on almost 20\% of frames. The prompt strategy is not as impressive, as it only exceeds Edge-Only or is comparable to it 78\% of the time and consistently underperforms the other three methods, though it updates the model more frequently. Notably, Shoggoth with adaptive frame sampling achieved better mAP than AMS with all frames 73\% of the time. Adaptive frame sampling prevents overfitting to recent frames, making the system respond to scene changes more effectively.

\begin{table}
\caption{Sensitivity to different sampling rates.}
\label{table:table3}
\centering
\vskip -8pt
\setlength\tabcolsep{4pt}
\begin{tabular}{c | c c c c c c c}
\toprule
$rate \rightarrow$ & 0.1 & 0.2 & 0.4 & 0.8 & 1.6 & 2.0 & Adaptive \\
\midrule
\textbf{Up BW (Kbps)} & 19 & 36 & 61 & 122 & 249 & 307 & 135 \\
\textbf{Average IoU} & 0.483 & 0.524 & 0.556 & 0.623 & 0.612 & 0.597 & 0.640 \\
\bottomrule
\end{tabular}
\vskip -9pt
\end{table}

\section{Conclusion}   \label{Conclusion}
In this paper, we introduce Shoggoth, a novel edge-cloud collaborative real-time video inference architecture. By adaptive online learning, Shoggoth maximizes the benefit of collaboration between edge and cloud resources to improve inference performance effectively. In addition, Shoggoth designs adaptive frame sampling, which significantly increases the robustness of scene changes and reduces communication overhead. Shoggoth outperforms state-of-the-art solutions in the trade-off between low latency and high accuracy, achieving 15\%–20\% accuracy improvement compared to the edge-only strategy and requiring 24× less uplink bandwidth to achieve similar accuracy to the cloud-only strategy.

\section*{Acknowledgment}
This work is supported by the Key Research and Development Program of Guangdong Province (Grant No. 2021B0101400003), the Creative Research Group Project of the NSFC (Grant No. 61821003), and the National Natural Science Foundation of China (Grant No. 62072196).

\bibliographystyle{IEEEtran}
\bibliography{IEEEabrv,references}

\end{document}